\documentclass[10pt,twocolumn,letterpaper]{article}

\usepackage[pagenumbers]{cvpr} 

\usepackage[dvipsnames,table]{xcolor}
\usepackage{multirow}
\usepackage{makecell}

\renewcommand{\paragraph}[1]{\vspace{1.25mm}\noindent\textbf{#1}}

\definecolor{cvprblue}{rgb}{0.21,0.49,0.74}
\usepackage[pagebackref,breaklinks,colorlinks,citecolor=cvprblue]{hyperref}

\def\paperID{5435} 
\def\confName{CVPR}
\def\confYear{2024}

\title{RepViT-SAM: Towards Real-Time Segmenting Anything}

\author{Ao Wang$^{1}$ \quad Hui Chen$^{1}$\thanks{Corresponding author.} \quad Zijia Lin$^{1}$ \quad Jungong Han$^{2}$ \quad Guiguang Ding$^{1}$\footnotemark[1] \\
		$^1$Tsinghua University \quad $^2$ The University of Sheffield \\
		{\tt\small wa22@mails.tsinghua.edu.cn \quad huichen@mail.tsinghua.edu.cn \quad linzijia07@tsinghua.org.cn} \\ 
		{\tt\small jungonghan77@gmail.com \quad dinggg@tsinghua.edu.cn }}

\begin{document}
\maketitle

\begin{abstract}
  Segment Anything Model (SAM)~\cite{kirillov2023segment} has shown impressive zero-shot transfer performance for various computer vision tasks recently~\cite{cao2023segment,ji2023segment,rajivc2023segment,zhao2023fast,xiong2023efficientsam}. However, its heavy computation costs remain daunting for practical applications. MobileSAM~\cite{zhang2023faster} proposes to replace the heavyweight image encoder in SAM with TinyViT~\cite{zhang2023faster} by employing distillation, which results in a significant reduction in computational requirements. However, its deployment on resource-constrained mobile devices still encounters challenges due to the substantial memory and computational overhead caused by self-attention mechanisms. Recently, RepViT~\cite{wang2023repvit} achieves the state-of-the-art performance and latency trade-off on mobile devices by incorporating efficient architectural designs of ViTs into CNNs. Here, to achieve real-time segmenting anything on mobile devices, following~\cite{zhang2023faster}, we replace the heavyweight image encoder in SAM with RepViT model, ending up with the RepViT-SAM model. Extensive experiments show that RepViT-SAM can enjoy significantly better zero-shot transfer capability than MobileSAM~\cite{zhang2023faster}, along with nearly $10\times$ faster inference speed. The code and models are available at \url{https://github.com/THU-MIG/RepViT}.

\end{abstract}

\section{Methodology}

SAM~\cite{kirillov2023segment} is composed of a heavyweight ViT-based image encoder and a lightweight prompt-guided mask decoder. Its huge image encoder accounts for the majority of the inference time overhead. Therefore, MobileSAM~\cite{zhang2023faster} suggests to replace the default ViT-H~\cite{dosovitskiy2020image} image encoder in SAM with a lightweight one, \ie, TinyViT~\cite{wu2022tinyvit}. TinyViT consists of four stages which gradually reduce the resolution. The initial stage of TinyViT is composed of convolution blocks utilizing inverted residual blocks~\cite{sandler2018mobilenetv2}. To downsample the resolution at the outset of the model, two convolution blocks with a stride of 2 are employed. Similarly, the convolution blocks with a stride of 2 is adopted to perform spatial downsampling between adjacent stages. In order to align the final resolution of TinyViT with that of the ViT-H image encoder of the original SAM, MobileSAM sets the stride of the last downsampling convolution in the TinyViT to 1. Besides, MobileSAM presents the decoupled distillation strategy to efficiently train the lightweight image encoder, in which the TinyViT model is directly distilled from the ViT-H in the original SAM without the prompt-guided mask decoder. Despite significantly reducing the computational requirements of segmenting anything, the deployment of MobileSAM on mobile devices still poses considerable challenges. As shown in \Cref{tab:sam-latency}, MobileSAM fails to run on an iPhone 12 due to its substantial memory footprint. Besides, on a Macbook, its inference time for processing a single image is 494ms, indicating significant room for improvement.

\begin{table}[t]
  \centering
  \caption{
    Comparison between RepViT-SAM and others in terms of latency. The latency (ms) is measured with the standard resolution~\cite{ghiasi2021simple} of 1024$\times$1024 on iPhone 12 and Macbook M1 Pro by Core ML Tools. OOM means out of memory.
  }
  \small
  \resizebox{\linewidth}{!}{
  \begin{tabular}{c|c|c|c|c}
  \toprule
   \multirow{2}{*}{Platform} &  \multicolumn{3}{c|}{Image encoder}  & \multirow{2}{*}{\shortstack{Mask\\ decoder}} \\
    \cmidrule{2-4}
    & \textbf{RepViT-SAM} & MobileSAM~\cite{zhang2023faster} & ViT-B-SAM~\cite{kirillov2023segment} \\
    \midrule
    iPhone & \textbf{48.9} & OOM & OOM & 11.6\\
    Macbook & \textbf{44.8} & 482.2 & 6249.5 & 11.8 \\
  \bottomrule
  \end{tabular}
  }
  \label{tab:sam-latency}
\end{table}

Recently, RepViT~\cite{wang2023repvit} demonstrates the state-of-the-art performance and latency trade-off on mobile devices by revisiting the efficient designs of CNNs from ViT perspective. RepViT employs the early convolutions~\cite{xiao2021early} as the stem, \ie, two convolutions with a stride of 2 to perform 4$\times$ downsampling for the input. It adopts the RepViT block, which is composed of structural reparameterized depthwise convolutions~\cite{chu2022make,ding2021repvgg} and feed forward module. A deep downsample module is employed between adjacent stages, which leverage a depthwise convolution with a stride of 2 and a pointwise convolution to perform spatial downsampling and channel dimension modulation, respectively. Besides, the squeeze-and-excitation~\cite{hu2018squeeze} layers is utilized for all stages in a cross-block manner. RepViT shows substantial advantages in terms of latency in high-resolution vision tasks~\cite{wang2023repvit}, due to its pure convolutional architecture. As shown in \Cref{tab:sam-latency}, after replacing the ViT-H image encoder with the RepViT-M2.3 model, RepViT-SAM exhibits a significant reduction in latency compared with others. On iPhone 12, RepViT-SAM can perform model inference smoothly. Besides, on Macbook, RepViT-SAM is nearly 10$\times$ faster than MobileSAM.

Following~\cite{zhang2023faster}, we train the RepViT-SAM by directly distilling the image encoder, \ie, RepViT-M2.3, from the ViT-H in the original SAM~\cite{kirillov2023segment}, leveraging a simple MSE loss. Like~\cite{zhang2023faster}, the stride of 2 in the last downsampling depthwise convolution in RepViT is set to 1 for making the output resolution compatible with the prompt-guided mask decoder in the original SAM~\cite{kirillov2023segment}.

\section{Experiments}

\subsection{Implementation details}
RepViT-SAM is trained for 8 epochs under the same setting as~\cite{zhang2023faster}. Like MobileSAM~\cite{zhang2023faster}, we use only 1\% data in the SAM-1B dataset~\cite{kirillov2023segment}. To expedite the training process, we precompute and save the image embeddings from the ViT-H image encoder prior to the distillation phase, which eliminates the need to run the forward process of ViT-H during distillation, like~\cite{zhang2023faster}. We evaluate the performance of RepViT-SAM on zero-shot edge detection on BSDS500~\cite{martin2001database,arbelaez2010contour}, zero-shot instance segmentation using COCO~\cite{lin2014microsoft}, and segmentation in the wild benchmark~\cite{zou2023generalized} (SegInW), zero-shot video object/instance segmentation using DAVIS 2017~\cite{pont20172017}/UVO v1.0~\cite{wang2021unidentified}, zero-shot salient object segmentation using DUTS~\cite{wang2017learning}, and zero-shot anomaly detection using MVTec-AD~\cite{bergmann2019mvtec}, following~\cite{kirillov2023segment,ke2023segment,rajivc2023segment,ji2023segment,cao2023segment}. 

\begin{table}[t]
  \centering
  \caption{
      Comparison results on zero-shot edge detection. Bold indicates the best, and underline indicates the second best.
  }
  \small
  \begin{tabular}{c|ccc}
  \toprule
  \multirow{2}{*}{Model} & \multicolumn{3}{c}{zero-shot edge detection} \\
      \cmidrule{2-4} 
      & ODS & OIS & AP\\
      \midrule
      ViT-H-SAM~\cite{kirillov2023segment}  & \textbf{.768} & \textbf{.786} & \textbf{.794} \\
      ViT-B-SAM~\cite{kirillov2023segment} & .743 & .764 & .726  \\
      MobileSAM~\cite{zhang2023faster} & .756 & .768 & .746  \\
      \rowcolor[gray]{0.92}
      RepViT-SAM & \underline{.764} & \textbf{.786} & \underline{.773} \\
  \bottomrule
  \end{tabular}
  \label{tab:sam-edge}
\end{table}

\subsection{Zero-shot edge detection}
Following~\cite{kirillov2023segment}, we utilize a 16$\times$16 regular grid of foreground points as prompts for the model. We apply NMS to remove redundant masks. Additionally, a Sobel filter is employed to the unthreshold probability maps of the masks, along with standard lightweight postprocessing, including edge NMS, as described in~\cite{kirillov2023segment}. As shown in~\Cref{tab:sam-edge}, our RepViT-SAM outperforms MobileSAM and ViT-B-SAM on all metrics. Compared with ViT-H-SAM, which is the largest SAM model with over 615M parameters, our small RepViT-SAM can obtain comparable performance in terms of ODS and OIS.  

\subsection{Zero-shot instance segmentation}
We leverage the state-of-the-art H-Deformable-DETR~\cite{jia2022detrs} with Swin-L~\cite{liu2021swin} as the object detector and prompt the model with its output boxes, like~\cite{ke2023segment,kirillov2023segment}. As shown in \Cref{tab:sam-instance}, our RepViT-SAM significantly outperforms MobileSAM and ViT-B-SAM by 1.7 and 1.9 AP, respectively. Besides, it can be observed that the models obtained by the decoupled distillation strategy, \ie, MobileSAM and RepViT-SAM, typically outperform the model obtained by the coupled distillation, \ie, ViT-B-SAM, on large objects, but exhibits comparatively poorer performance on small objects. Such phenomenon may be attributed to the fact that the decoupled distillation strategy enables more transfer of macro-level visual features but falls short in fine-grained features.

\subsection{Segmentation in the wild benchmark}
We follow~\cite{ke2023segment} to equip the Grounding-DINO~\cite{liu2023grounding} as box prompts to evaluate the model on the zero-shot track. As shown in~\Cref{tab:sam-instance}, our RepViT-SAM outperforms MobileSAM and ViT-B-SAM by 2.2 and 1.3 mean AP, respectively. It well indicates the strong transferability of RepViT-SAM to diverse downstream segmentation tasks.

\begin{table}[t]
  \centering
  \caption{
      Comparison results on zero-shot instance segmentation and segmentation in the wild benchmark (SegInW). Bold indicates the best, and underline indicates the second best.
  }
  \small
  \resizebox{\linewidth}{!}{
  \begin{tabular}{c|cccc|c}
  \toprule
  \multirow{2}{*}{Model} & \multicolumn{4}{c|}{zero-shot instance segmentation} & SegInW \\
      \cmidrule{2-6} 
      & AP & AP$^{S}$ & AP$^{M}$ & AP$^{L}$ & Mean AP\\
      \midrule
      ViT-H-SAM~\cite{kirillov2023segment}  & \textbf{46.8} & \textbf{31.8} & \textbf{51.0} & \textbf{63.6} & \textbf{48.7} \\
      ViT-B-SAM~\cite{kirillov2023segment} & 42.5 & \underline{29.8} & 47.0 & 56.8 & 44.8 \\
      MobileSAM~\cite{zhang2023faster} & 42.7 & 27.0 & 46.5 & 61.1 & 43.9 \\
      \rowcolor[gray]{0.92}
      RepViT-SAM &  \underline{44.4} & 29.1 & \underline{48.6} & \underline{61.4} & \underline{46.1} \\
  \bottomrule
  \end{tabular}
  }
  \label{tab:sam-instance}
\end{table}

\begin{table}[t]
  \centering
  \caption{
      Comparison results on zero-shot video object segmentation (z.s. VOS) and video instance segmentation (z.s. VIS). Bold indicates the best, and underline indicates the second best.
  }
  \small
  \begin{tabular}{c|ccc|c}
  \toprule
  \multirow{2}{*}{Model} & \multicolumn{3}{c|}{z.s. VOS} & z.s. VIS \\
      \cmidrule{2-5} 
      & \textbf{$\mathcal{J\&F}$} & \textbf{$\mathcal{J}$} & \textbf{$\mathcal{F}$} & AR100\\
      \midrule
      ViT-H-SAM~\cite{kirillov2023segment}  & \textbf{77.4} & \textbf{74.6} & \textbf{80.2} & \textbf{28.8} \\
      ViT-B-SAM~\cite{kirillov2023segment} & 71.3 & 68.5 & 74.1 & 19.1 \\
      MobileSAM~\cite{zhang2023faster} & 71.1 & 68.6 & 73.6 & 22.7 \\
      \rowcolor[gray]{0.92}
      RepViT-SAM &  \underline{73.5} & \underline{71.0} & \underline{76.1} & \underline{25.3} \\
  \bottomrule
  \end{tabular}
  \label{tab:sam-video}
\end{table}

\subsection{Zero-shot video object/instance segmentation}
We leverage SAM-PT~\cite{rajivc2023segment} to evaluate the performance of RepViT-SAM on the validation splits of DAVIS 2017~\cite{pont20172017} and UVO~\cite{wang2021unidentified} datasets. We simply replace the original SAM in SAM-PT with our RepViT-SAM, using the CoTracker~\cite{karaev2023cotracker} as the point tracker without the reinitialization strategy. As shown in \Cref{tab:sam-video}, on DAVIS 2017, our RepViT-SAM obtains a mean \textbf{$\mathcal{J\&F}$} score of 73.5, significantly outperforming MobileSAM and ViT-B-SAM by 2.4 and 2.2, respectively. Besides, on UVO, RepViT-SAM surpasses MobileSAM and ViT-B-SAM with considerable margins of 2.6 and 6.2, respectively. These results show the superiority of RepViT-SAM compared with other lightweight ones is not limited to image segmentation tasks but can also be extended to video segmentation tasks.

\begin{table}[t]
  \centering
  \caption{
      Comparison results on zero-shot salient object segmentation (z.s. s.o.s.) and zero-shot anomaly detection (z.s. a.d.). Bold indicates the best, and underline indicates the second best.
  }
  \small
  \begin{tabular}{c|c|c}
  \toprule
  \multirow{2}{*}{Model} & z.s. s.o.s. & z.s. a.d. \\
      \cmidrule{2-3} 
      & \textbf{$\mathcal{M}$ $\downarrow$} & \textbf{$\mathcal{F}_{p}$} \\
      \midrule
      ViT-H-SAM~\cite{kirillov2023segment}  & \textbf{0.046} & \underline{37.65} \\
      ViT-B-SAM~\cite{kirillov2023segment} & 0.121 & 36.62 \\
      MobileSAM~\cite{zhang2023faster} & 0.147 & 36.44 \\
      \rowcolor[gray]{0.92}
      RepViT-SAM &  \underline{0.066} & \textbf{37.96} \\
  \bottomrule
  \end{tabular}
  \label{tab:sam-salient}
\end{table}

\subsection{Zero-shot salient object segmentation}
We follow~\cite{ji2023segment} to first leverage RepViT-SAM to generate N potential objects masks for a given image. Subsequently, we select the most appropriate mask by evaluating its alignment with the ground truth. The intersection over union (IoU) score is adopted as the alignment metric. Following~\cite{ji2023segment}, we report mean absolute error (MAE) score. Note that a lower MAE indicates better segmentation results. As shown in \Cref{tab:sam-salient}, our RepViT-SAM shows its superiority compared with MobileSAM and ViT-B-SAM in terms of MAE score.

\subsection{Zero-shot anomaly detection}
We leverage SAA+~\cite{cao2023segment} to evaluate the performance of RepViT-SAM on the MVTec-AD benchmark dataset. We simply replace the original SAM in SAA+ with our RepViT-SAM. We follow~\cite{cao2023segment} to report max-F1-pixel (\textbf{$\mathcal{F}_{p}$}), which quantifies the F1-score for pixel-wise segmentation at the optimal threshold. As shown in \Cref{tab:sam-salient}, our RepViT-SAM achieves the best performance in terms of \textbf{$\mathcal{F}_{p}$}, even outperforming ViT-H-SAM. It indicates that for certain specific downstream task, the decoupled distilled model can achieve performance than the original SAM, along with much lower latency, showing the promising applications of RepViT-SAM.

\section{Visualization}
\begin{figure*}[t]
\centering
    \includegraphics[width=0.9\linewidth]{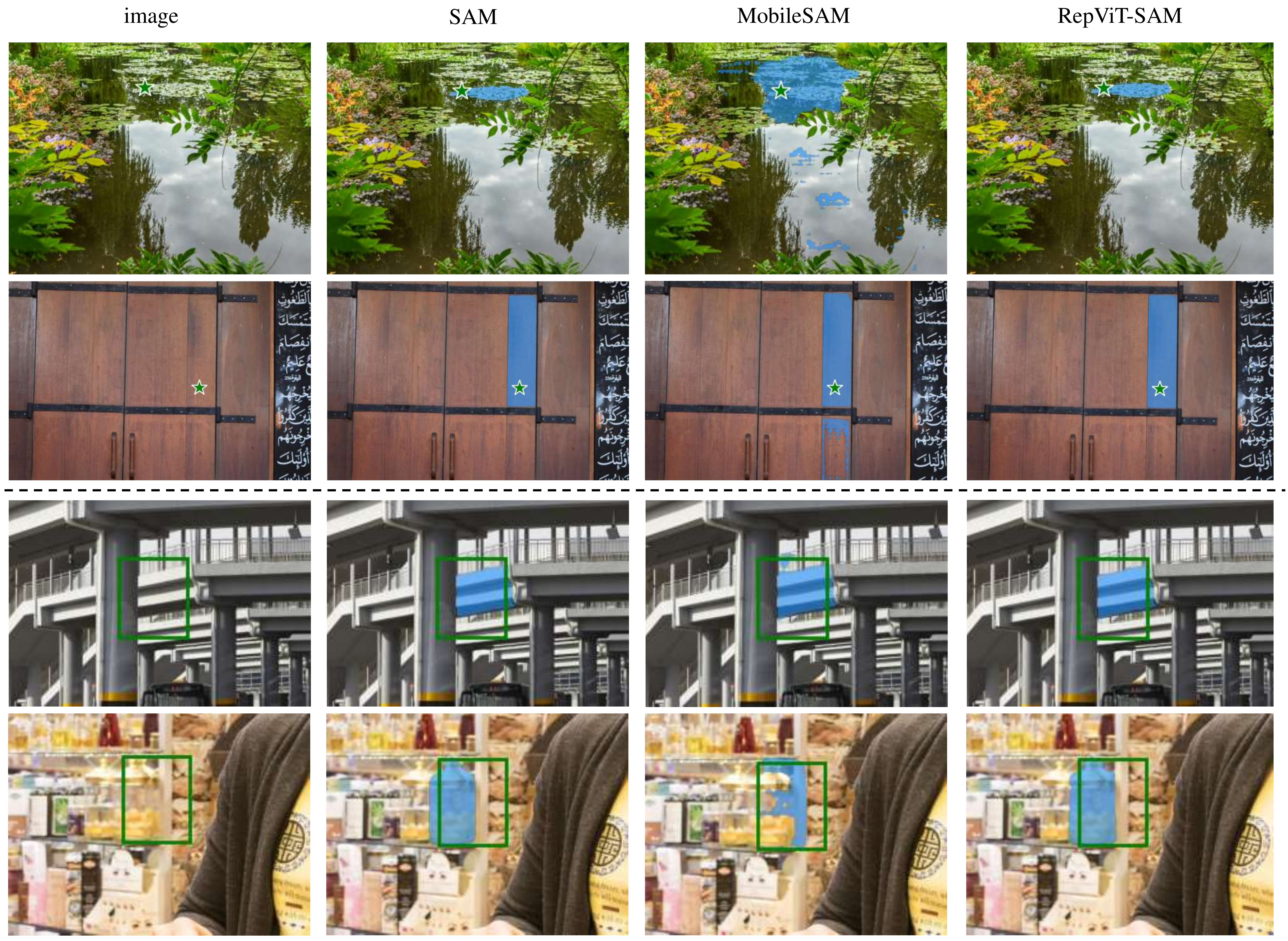}
    \caption{Mask predictions of SAM, MobileSAM, and RepViT-SAM with point prompts (top) and box prompts (bottom).}
    \label{fig:sam}
\end{figure*}

\begin{figure*}[t]
\centering
    \includegraphics[width=0.9\linewidth]{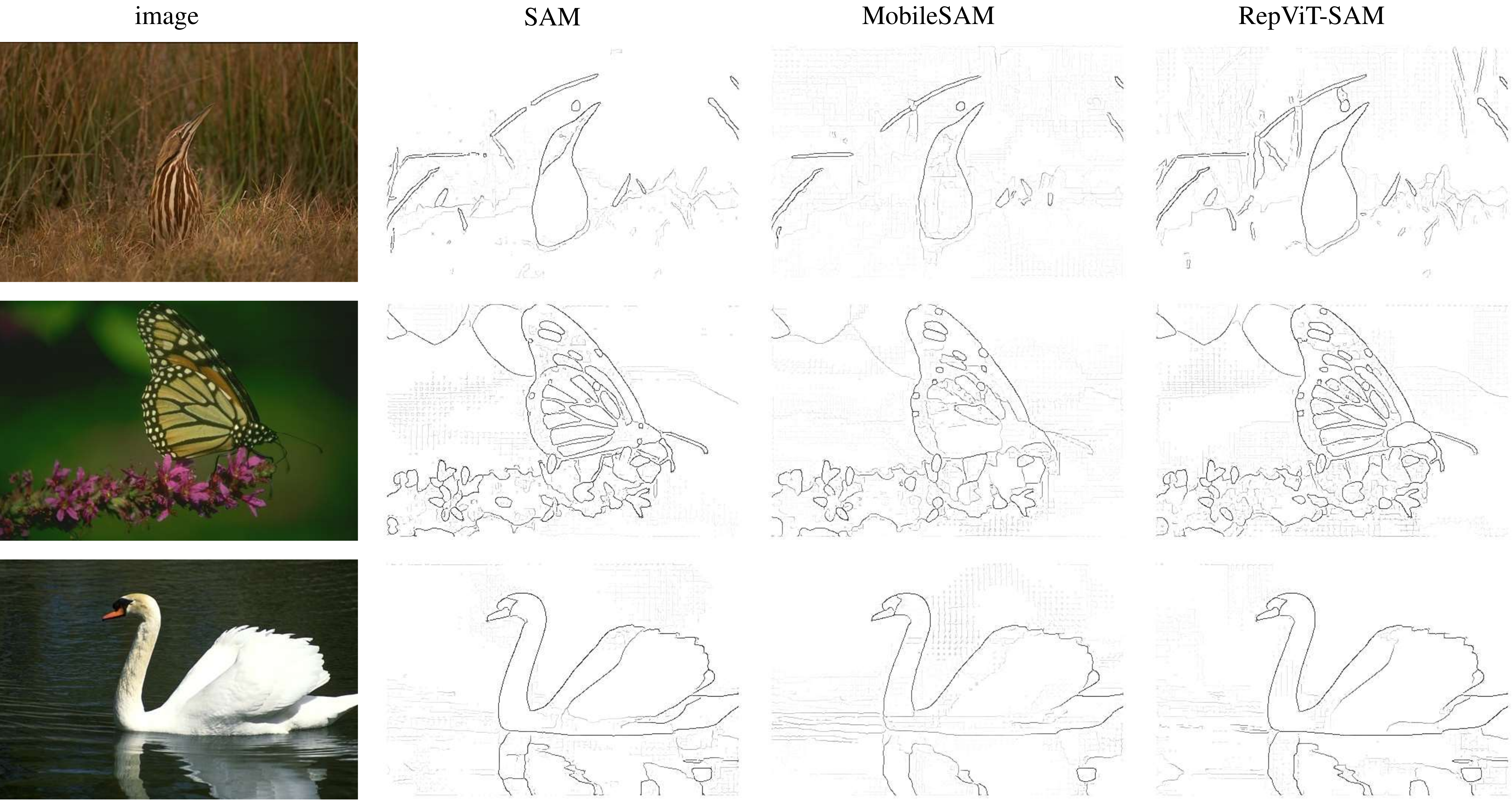}
    \caption{Visualization results of SAM, MobileSAM, and RepViT-SAM for zero-shot edge detection on BSDS500.}
    \label{fig:edge}
    \vspace{-3pt}
\end{figure*}

In Figure \ref{fig:sam}, we showcase the predicted masks of SAM \cite{kirillov2023segment}, MobileSAM \cite{zhang2023faster}, and RepViT-SAM with point and box prompts. It shows that RepViT-SAM can generate high-quality mask predictions similar to SAM across various scenarios, where MobileSAM may encounter challenges in accurately predicting masks. Furthermore, we present qualitative comparison results for zero-shot edge detection in Figure \ref{fig:edge}. RepViT-SAM effectively produces reasonable edge maps, achieving performance comparable to SAM. In contrast, MobileSAM may struggle to generate accurate edge maps in regions that contain intricate details.

\section{Conclusion}
In summary, our RepViT-SAM demonstrates outstanding efficiency while maintaining impressive transfer performance for various downstream tasks. We hope that our RepViT-SAM model can serve as a robust baseline for SAM in edge deployments, showcasing its potential for practical applications.

{
    \small
    \bibliographystyle{ieeenat_fullname}
    \bibliography{main}
}


\end{document}